\documentclass{mva_style}
\usepackage{graphicx}
\usepackage{amsfonts}
\usepackage{amsmath}
\usepackage{bm}
\usepackage[bottom]{footmisc}

\finalcopy 

\begin{document}
\title{Visual Rhythm Prediction with Feature-Aligning Network}

\author{
  Yutong Xie\\
  Shanghai Jiao Tong University\\
  {\tt xxxxxyt@sjtu.edu.cn}\\
  \and
  Haiyang Wang\\
  Shanghai Jiao Tong University\\
  {\tt wanghaiyang@sjtu.edu.cn}\\
  \and
  Yan Hao\\
  Shanghai Jiao Tong University\\
  {\tt honeyhaoyan@sjtu.edu.cn}\\
  \and
  Zihao Xu\\
  Shanghai Jiao Tong University\\
  {\tt shsjxzh@sjtu.edu.cn}\\
}

\maketitle

\section*{\centering Abstract}
\textit{
    In this paper, we propose a data-driven visual rhythm prediction method, which overcomes the previous works' deficiency that predictions are made primarily by human-crafted hard rules. In our approach, we first extract features including original frames and their residuals, optical flow, scene change, and body pose. These visual features will be next taken into an end-to-end neural network as inputs.  Here we observe that there are some slight misaligning between features over the timeline and assume that this is due to the distinctions between how different features are computed. To solve this problem, the extracted features are aligned by an elaborately designed layer, which can also be applied to other models suffering from mismatched features, and boost performance. Then these aligned features are fed into sequence labeling layers implemented with BiLSTM \cite{bilstm} and CRF \cite{crf} to predict the onsets. Due to the lack of existing public training and evaluation set, we experiment on a dataset constructed by ourselves based on professionally edited Music Videos (MVs), and the F1 score of our approach reaches 79.6.
}

\section{Introduction}

Visual rhythm prediction has caught people's attention for years, since it can enable many valuable applications like automated video editing. 
This problem can be described as given a segment of videos, we want to decide whether each time point is an onset or not, as earlier works discussed \cite{obama}\cite{chu2012rhythm}. But most of the previously proposed methods have a main disadvantage or disability: they primarily rely on human-crafted hard rules to compute the visual onsets \cite{obama}\cite{arguello2016exploring}\cite{chu2012rhythm}\cite{chen2011visual}, and can only perform well on a small set of specific videos, like dancing video without camera moving.

We admit that, the visual rhythm is hard to rigidly define with a simple formula, due to the variety and complexity of rhythm-related visual cues, including: 1. visual content changes; 2. movement of lens or camera; 3. environmental lighting conversion; 4. scene changes; 5. motion of performers, etc.

However, the visual rhythm can be indirectly reflected by the corresponding musical rhythm to some extent, especially in the professionally edited MVs, which enables us to learn how to predict visual onsets from MVs of high quality. So we propose a data-driven method for visual rhythm prediction with an end-to-end \emph{Feature-Aligned Network} (FAN), and train it with sufficient data.

\begin{figure}[t]
  \begin{center}
    \includegraphics[width=0.5\textwidth]{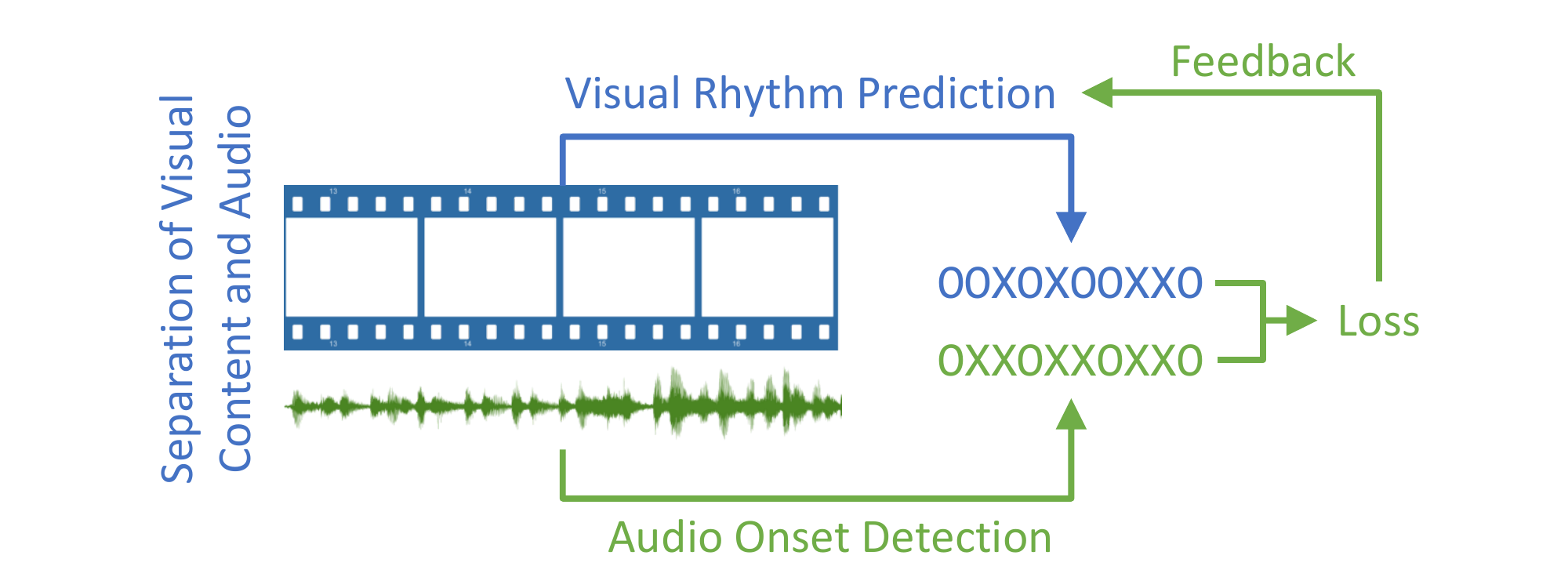}
  \end{center}
  \caption{
    For a given video, we can predict the visual rhythm by deciding a time point to be an onset or not. While training the predictor with professionally edited MVs, the results of audio onset detection will be used as labels to rectify the predictor. The blue parts will be executed at both testing and training stage, while the green parts will only be executed at the training stage.
  }
  \label{fig:overview}
\end{figure}

In our method,  the visual rhythm is related with various visual cues mentioned above, we first extract the features closely linked with these cues, including 1. original frames; 2. frame residuals; 3. optical flow; 4. scene change; 5. body pose. The detailed reason why and by how we extract them will be further explained in Section \ref{sec:visual-feature-extraction}.

Then these extracted visual features are fed into FAN to predict the onsets. Here, we observe that there are some slight misaligning between features over the timeline, and assume that this is due to the distinctions between how different features are computed. For example, the body pose is decided by only a single frame, while the frame residuals depend on two consecutive frames and scene change is influenced by a few continuous frames. So after the extraction and transformation, we align features with an elaborately designed layer, which can also be applied to other models suffering from mismatched features, and bring in performance improvement. Next, the aligned features are fed into final layers implemented with BiLSTM \cite{bilstm} and CRF \cite{crf} to predict the onsets, as we further formalize the visual rhythm prediction as the general sequence labeling problem. The architecture and details will be further explained in Section \ref{sec:our-approach}.

Due to the lack of public training and evaluation set, we construct a \emph{MV Visual Rhythm} (MVVR) dataset by ourselves based on MVs published on the Internet. In our dataset, as we assume that the visual and musical rhythm will properly match each other in professionally edited MVs, the results of musical onset detection is taken as ground truth. In the experiment, the F1 score reaches $79.6$, which proves our method to be effective.

\balance

\section{Related Work}

\subsection{Visual rhythm prediction} 
In this part, we will briefly review the previous work on visual rhythm prediction. Davis et al. \cite{obama} suggest that the sudden visible deceleration of the moving object can indicate the visual rhythm, and measure it by calculating the optical flow of videos. However, this rule-based method cannot distinguish the motion of central subject from the background or camera motion, and even minor camera motion disturbance can be a great interference for the detection,  the motion patterns are too complicated to be well described with concise formulas.
 
Arg{\"u}ello et al. \cite{arguello2016exploring}, Chu et al. \cite{chu2012rhythm} in another perspective, come up with the similar idea that visual rhythm can be treated as periodic patterns in actual motion, and then employ different motion detection methods to eventually extract visual rhythm onsets.

Chen et al. \cite{chen2011visual} describe visual rhythm in a more rough way as the occurrence frequency of rhythmic events like human movement and environment lighting change. Absolute frame difference and 2D angle-magnitude histogram of optical flows are used in this article to measure such frequency. However, it can not precisely predict the visual onset events timing and only estimate how intense the visual rhythm is.

The previously mentioned methods are all rule-based, and their key problem lies in that it can only perform well on a small set of specific videos, like dancing video without camera moving. Therefore, we refer to deep representation learning to handle more complicated patterns and enrich the expressiveness of our model.


\subsection{Musical onset detection}

Musical onset detection is an area that has been explored for years and can bring us much inspiration, since it is also modeled as a sequence labeling problem. Bello et al. \cite{bello2004use}  present a rule-based model 
and have achieved one of the best results by unsupervised methods. But this has soon been outperformed by methods employing deep neural networks \cite{gong2018towards}\cite{schluter2014improved}\cite{eyben2010universal}, which convinces us to address the visual rhythm prediction problem with deep learning methods.

\begin{figure}[h]
  \begin{center}
    \includegraphics[width=0.5\textwidth]{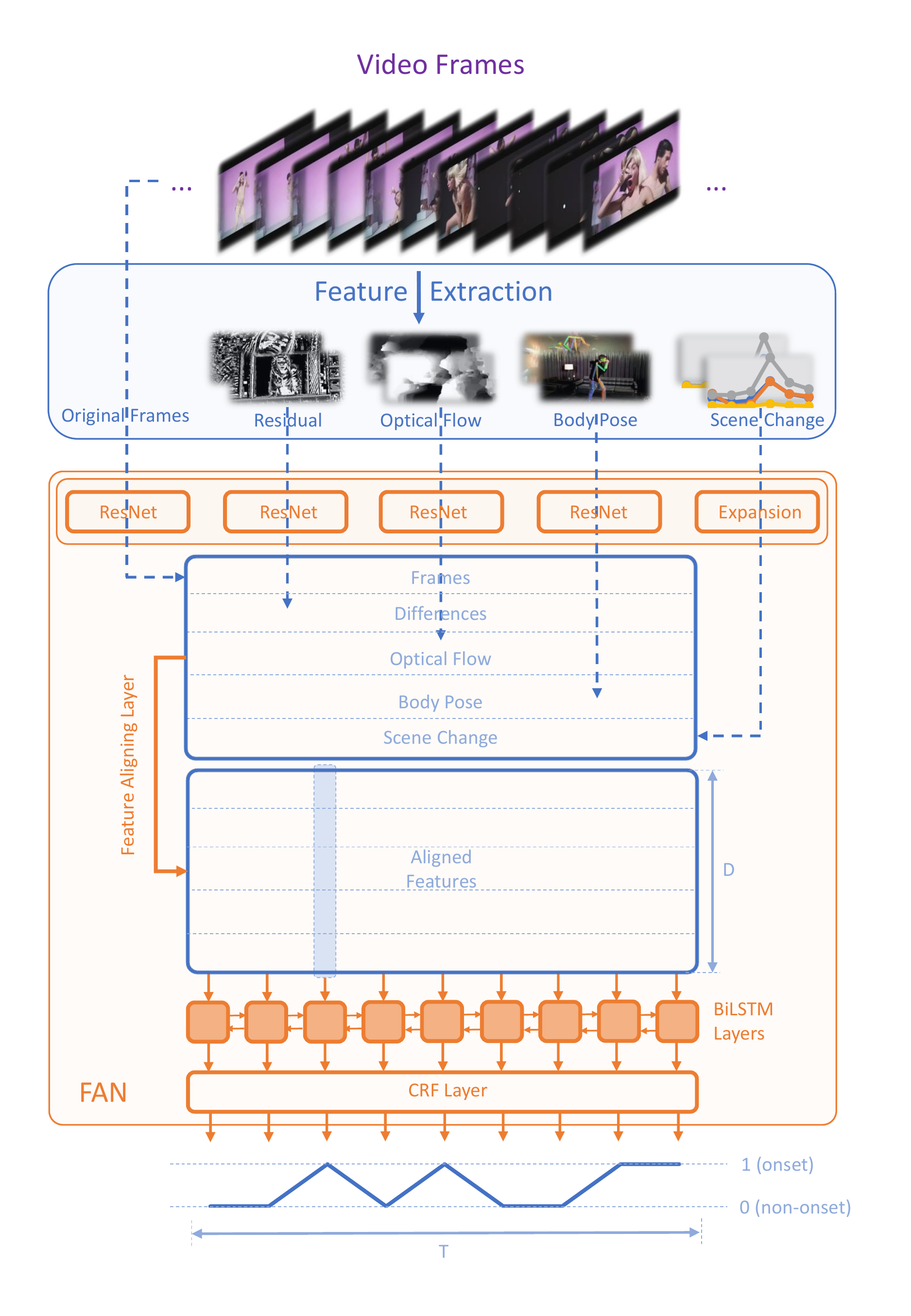}
  \end{center}
  \caption{
    Our approach with the end-to-end Feature-Aligning Network (FAN). Extracted visual features are fed into the sequence labeling layers after transformation and aligning, which predicts the rhythm onsets.
  }
  \label{fig:network}
\end{figure}

\section{Visual Feature Extraction}
\label{sec:visual-feature-extraction}

In our method, we first extract rhythm-related features from the visual content, and the results will be subsequently fed into the end-to-end network --- FAN, to predict visual onsets.

Considering the visual content changes, movement of lens or camera, environmental lighting conversion, scene changes and motion of performers as visual rhythm cues in videos, we extract features as following:

\begin{description}
\item[Original Frames] From original frames, we can know what is shown in the video, which is helpful to the rhythm prediction when the video is periodically changing the performing content. Here we simply take frames from RGB channels.

\item[Frame Residuals] The residual of frames implies the movement of the lens or objects, which can also reflects the visual rhythm \cite{chen2011visual}. In this part, we directly compute the residual between two adjacent frames with subtraction.

\item[Optical Flow Detection] Optical flow is the pattern of apparent motion of image objects between two consecutive frames caused by the movement of objects or the camera. It reflects the intensity of the action in videos, which offers us some visual rhythm cues. In this part, we adapt the Lucas-Kanade method \cite{optical_flow}, which solves the basic optical flow equations for pixels in a neighbourhood by the least squares criterion, to gain a feature map whose size is the same as original frames.

\item[Scene Change Detection] Scene change detection divides a video into physical shots, which are video sequences that consists of continuous number of video frames for particular action. In this part, we make use of a video sequence detection method \cite{scene_change} based on the three dimensional histogram of color images, which gets the pattern of scene change by comparing the difference of histograms and their size between consecutive frames.

\item[Body Pose Detection] Most musical videos contain human movement (In the searching results of ``MV'' on YouTube, 99\% of the videos contain human figures and their body motion), and human's regular move pattern is also a good indicator for visual rhythm \cite{arguello2016exploring}\cite{chu2012rhythm}.  For this part, we refer to the state-of-art work of pose estimation \cite{RMPE} to extract body movement in a key point level.
\end{description}

\section{FAN: Feature-Aligning Network}
\label{sec:our-approach}

In the end-to-end Feature-Aligning Network, we take the previously extracted features as inputs, and first transform them into a common feature vector space (Section \ref{sec:feature-transformation}). After this, to alleviate the observed mismatching problem, we align them with an elaborately designed layer (Section \ref{sec:aligning-layer}). These aligned features are next fed into sequence labeling layers (Section \ref{sec:sequence-labing-layers}) to predict the onsets. The whole prediction process is illustrated as Figure \ref{fig:network}.

\subsection{Feature Transformation}
\label{sec:feature-transformation}

For the original frames and feature maps extracted by frame residuals, optical flow detection and body pose detection, we further transform them into feature vectors with ResNet-34 \cite{resnet}.

As for the low-dimensional feature extracted by scene change detection, we expand it into a high-dimensional feature space with fully connected layers.

\subsection{Feature Aligning Layer}
\label{sec:aligning-layer}

\begin{figure}[h]
  \begin{center}
    \includegraphics[width=0.5\textwidth]{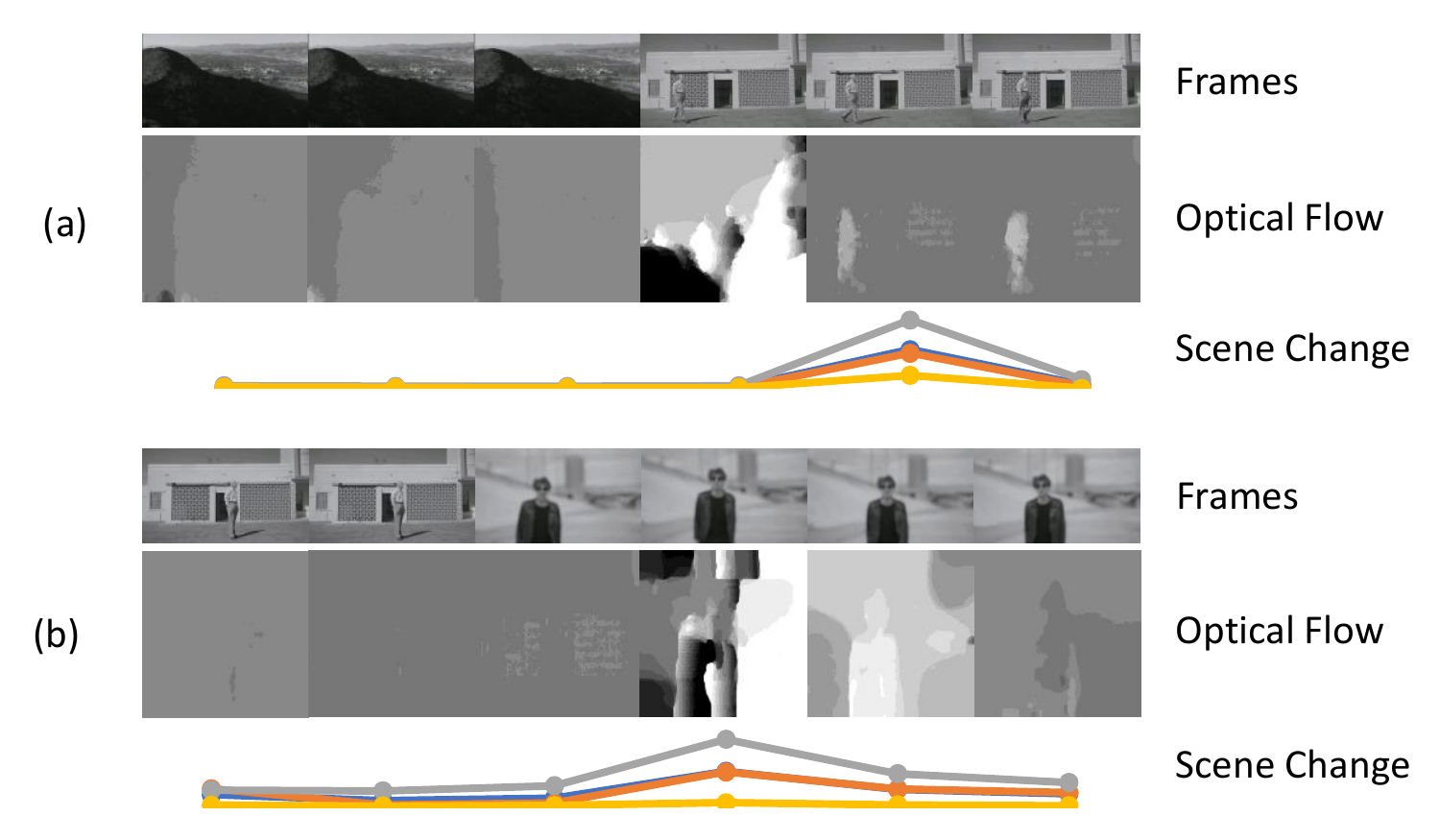}
  \end{center}
  \caption{
    Two segments drawn from one video. In (a), the optical flow exactly matches the change in the frames, but the scene change detection result seems to be delayed. In (b), ``peaks'' of the optical flow and scene change are in the same position, but both fall behind the original frames.
  }
  \label{fig:align-example}
\end{figure}

We observe that there are some slight misaligning between features over the timeline, as shown in Figure \ref{fig:align-example}.  Since the input video frames are selected by a common criteria, we assume that this problem is due to the distinctions between how different features are computed. Body pose is decided by only a single frame, corresponding to the original frame. As for the frame residuals and optical flow, they depend on two consecutive frames, while scene changes are influenced by a few continuous frames. So on a single time point, frames that are related with each feature vector are different, possibly leading to the misaligning problem. 

Besides, misaligned features can seriously harm the performance of the prediction network. This is because all features along a time point will be mixed and then mapped to a new feature space during the subsequent feature transformation, which turns the misaligned features to worthless even adverse noise. In addition, the offset is position-sensitive, which means that it cannot be eliminated by simply working on the process of feature extraction.

\begin{figure}[h]
  \begin{center}
    \includegraphics[width=0.5\textwidth]{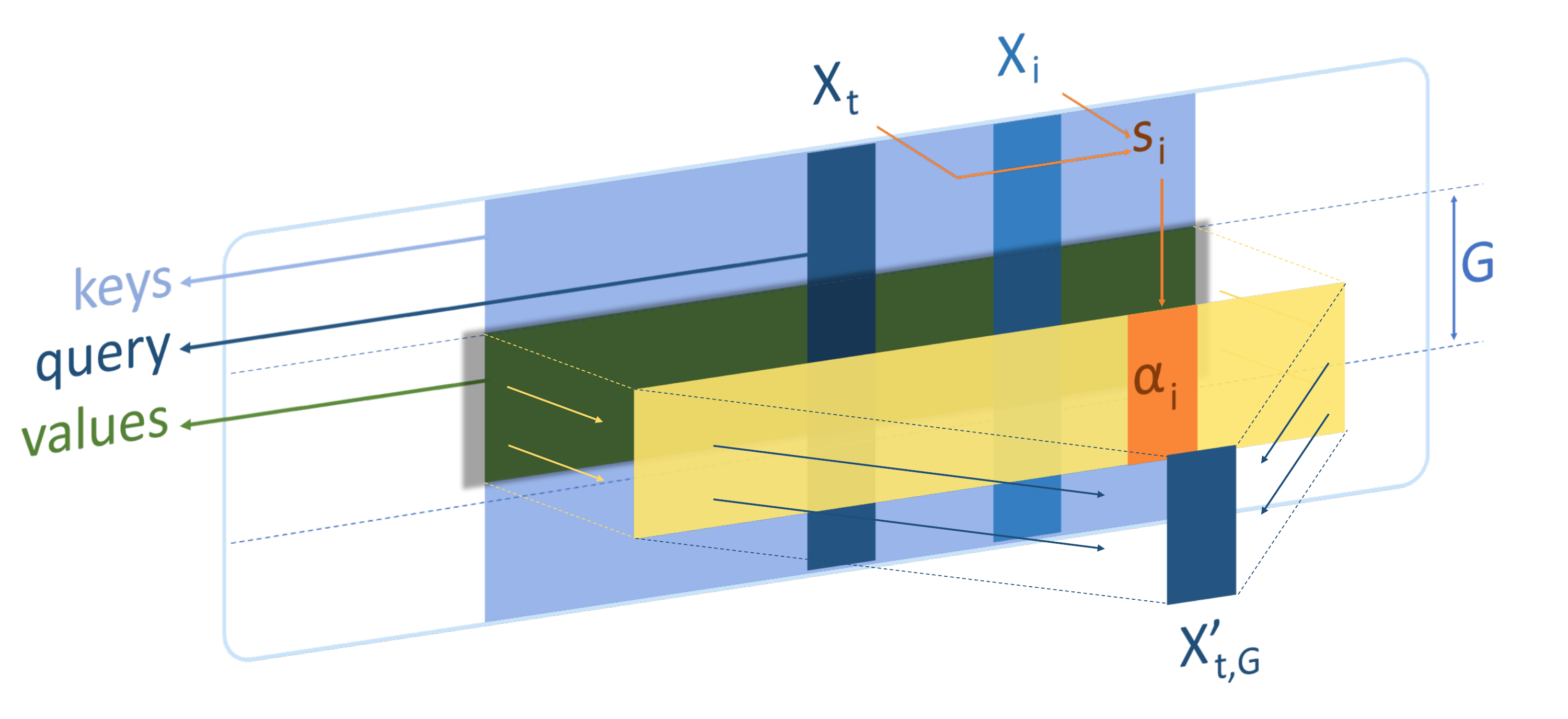}
  \end{center}
  \caption{
    The results of rearranging a group of features by feature aligning layer is the weighted sum of values (a range of time points with this feature group), with attention weights computed by scoring the relevance of the query (this time point with all feature) and keys (a range of time points with all features).
  }
  \label{fig:align-matrix}
\end{figure}

Therefore, we propose to alleviate this problem by a feature aligning layer, which can automatically learn how to align features over the timeline. More specifically, it employs the attention mechanism over sequence \cite{attention}, with which great progress has been made in Natural Language Processing (NLP) area, especially on the machine translation task. For a group of features $G$, the aligning layer rearrange it with the scaled dot-product attention as Figure \ref{fig:align-matrix}.

Denote $t=1,\dots,T$ as the time indicator with $T$ the length of this segment of video, $d\in G$ as the feature indicator of this group, $\bm{X}\in\mathbb{R}^{T\times D}$ as the concatenated extracted features with $D$ the total dimension, $p$ as the maximum offset. 

Then by applying two group-specific transformation matrices $\bm{W}^q,\bm{W}^K\in\mathbb{R}^{D\times D}$, we can define the scaled dot-product attention scores $\bm{s}\in\mathbb{R}^{2p+1}$ and the normalized attention weight $\bm{\alpha}\in[0,1]^{2p+1}$, from which the rearranged results $\bm{X}'\in\mathbb{R}^{T\times |G|}$ can be computed
as following

\begin{align}
\label{eq:weighted-sum}
s_i&=\frac{(\bm{W}^q\bm{X}_t)(\bm{W}^K\bm{X}_{i})^{\rm T}}{\sqrt{D}} \in\mathbb{R} \\
\alpha_i&=\frac{\exp(s_i)}{\sum_{j=t-p}^{t+p}\exp(s_j)} \in [0,1] \\
X_{t,d}'&=\sum_{i=t-p}^{t+p}\alpha_i X_{i,d} \in\mathbb{R} 
\end{align}

Here $\bm{X}_t$ plays the role of query, $\bm{X}_i$ plays the role of key, and $X_{t,d}$ plays the role of value. The result of rearranging a group of features by aligning layer is the weighted sum of values, with attention weights computed by scoring the relevance of the query and keys. Furthermore, by examining the attention weights, we can know how features are rearranged to align with each other.

The feature aligning layer is also an organic part of the end-to-end network, since the aligned results are next fed into subsequent layers, in which the training loss is back propagated. Thus the parameter matrices $\bm{W}^q,\bm{W}^K$, determining how values are weighted, can be optimized by taking gradient on the loss.

\subsection{Sequence Labeling Layers}
\label{sec:sequence-labing-layers}

Given a sequence of frames, deciding whether one of it is an onset or not is indeed a sequence labeling problem. To utilize the information of adjacent frames, we employ the Bidirectional Long Short Term Memory (BiLSTM) \cite{bilstm} modules instead of separately predicting on each time point.

Aside from considering the information adjacent frames, it is also beneficial to jointly predict over the whole sequence. For example, in a smoothing video, it is unlikely to have two consecutive onsets in a very short time. Thus we apply a final Conditional Random Filed (CRF) \cite{crf} layer to make the prediction aware of consecutive predictions.

\section{Experiment}

\subsection{Dataset}

To our best knowledge, there exists no public visual rhythm prediction dataset. So we construct our own dataset based on YouTube-music-video-5M\footnote{YouTube-music-video-5M: \\https://github.com/keunwoochoi/YouTube-music-video-5M}, a collection offers various styles of MVs, and published professionally edited MVs on YinYueTai\footnote{YinYueTai: http://www.yinyuetai.com/}, a popular music website. By manually filtering out poorly edited MVs, where the visual and musical rhythm don't match each other, we obtained a video set of size $800$. 

Fixing the segment length at $T=20$, for each piece of video segment, we first separate the visual and audio contents, then 

\begin{enumerate}
    \setlength{\itemsep}{-1ex}
    \item Extract video frames from RGB channels as input $\bm{X}\in\mathbb{R}^{T\times H\times W\times 3}$ where $H,W=224$ is the normalized height and width. The extraction is at $4$fps, to balance the labeling sensitivity and human's tolerance of visual rhythm deviation;
    \item Take the musical onset detection result as the ground truth $\bm{y}\in\{0,1\}^T$.
\end{enumerate}

Moreover, we wash out some excessively intense or smooth segments whose audio onset ratio (number of frames containing audio onsets divided by the segment length $T$ in the sense of $4$fps) beyond the range $[0.2,0.8]$.

Finally, we form an MV Visual Rhythm (MVVR) dataset whose statistic information is listed as Table \ref{tab:dataset}, where the counting is taken on musical onset detection results, i.e. the ground truth.

\begin{table}[h]
  \caption{MV Visual Rhythm Dataset}
  \begin{center}
    \begin{tabular}{c | c}
      \hline
      \hline
      & Counting \\
      \hline
      Total Frames & 850,620 \\
      Onset Labels & 318,932 \\
      Non-onset Labels & 531,688 \\
      \hline
      \hline
    \end{tabular}
    \label{tab:dataset}
  \end{center}
\end{table}

\subsection{Audio Onset Detection}
\label{sec:music-feature-extraction}

As for the audio onset detection, we follow the work of Sebastian B{\"o}ck et al. \cite{onset}, in which the spectral flux onset strength envelope is computed, and onset events are located by picking peaks in the envelope.

\subsection{Implementation Details}

In our implementation achieving the best performance, we employ ResNets of $34$ layers and $2$ layers fully connected layers to first transform extracted features, and the dimension of transformed feature space $D$ is $500$. Then in the aligning layer, the maximum offset $p$ is set as $2$, and features are grouped by the extraction. In the sequence labeling layers, we use $2$ layers of BiLSTMs and the CRF to predict the visual onsets, where the hidden state is of dimension $256$. During training, the Adam optimizer is utilized, and the learning rate is set as $3\times 10^{-5}$.

\subsection{Results and Analysis}

We estimate our prediction results with the metric of F1 score, in which the precision is defined as the true onset predictions count divided by all onset predictions count, and recall is defined as the true onset predictions count divided by all onsets count in the ground truth.

We experiment on the performance achieved by different features. Here all components in FAN are involved. The main results are shown in Table \ref{tab:feature}.

\begin{table}[h]
  \caption{Performance improvement with different visual features.}
  \begin{center}
    \begin{tabular}{l | c  c  c}
      \hline
      \hline
       & Precision & Recall & F1 Score \\
      \hline
      Frames+Residuals & 61.7 & \textbf{99.4} & 76.1 \\
      Optical Flow & 61.7 & 99.2 & 76.1 \\
      Scene Change & 56.3 & 88.1 & 68.7 \\
      Body Pose & 61.2 & 87.4 & 72.0 \\
      All Features & \textbf{78.2} & 81.0 & \textbf{79.6} \\
      \hline
      \hline
    \end{tabular}
    \label{tab:feature}
  \end{center}
\end{table}

As the above table shows, the combination of frames and residuals obtain the highest recall as $99.4$, probably because of the frequent object or camera motion which leads to the predictor's tending to label frames as visual onsets. Frame residuals and optical flow gain similar results, probably because they both measure the differences between two consecutive frames. However, scene change and body pose perform not as ideal as the former features. This is possibly because, scene changes and the body pose may be sparse in a small number of video segments, due to the variety of our videos, since not every frame contains changing of view or figures to be detected. Above all, the combination of all features reaches the best precision of $78.2$ and F1 score of $79.2$.

 \textbf{}
\section{Conclusion}

In this paper, by formalizing the visual rhythm prediction as a sequence labeling problem, we proposed a data-driven method with an end-to-end network named Feature-Aligning Network, which utilizes the visual features including original frames and their residuals, optical flow, scene change and body pose, to predict visual onsets with sequence labeling layers consisting of BiLSTM and CRF. Here, we observed the slight misaligning between features over the timeline, and assumed that this is due to the distinctions between how different features are computed. Then we addressed this problem with an elaborately designed aligning layer, which can also be applied to other models suffering from mismatched features, and bring in performance improvement. Lastly, we constructed a MV Visual Rhythm dataset based on professionally edited MVs to fill the vacancy of the training set and public evaluation set. In the experiment on this dataset, the F1 score of our approach reached $79.6$, which proved our method to be effective.

\end{document}